\documentclass{article}

\usepackage{PRIMEarxiv}

\usepackage[utf8]{inputenc} 
\usepackage[T1]{fontenc}    
\usepackage{hyperref}       
\usepackage{url}            
\usepackage{booktabs}       
\usepackage{amsfonts}       
\usepackage{nicefrac}       
\usepackage{microtype}      
\usepackage{lipsum}
\usepackage{fancyhdr}       
\usepackage{graphicx}       
\usepackage{multirow}
\usepackage{multicol}
\graphicspath{{media/}}     

\title{Tracking Fast by Learning Slow: An Event-based Speed Adaptive Hand Tracker Leveraging Knowledge in RGB Domain
}

\author{
  Chuanlin Lan, Ziyuan Yin, Arindam Basu, Rosa H. M. Chan\\
  Department of Electrical Engineering \\
  City University of Hong Kong \\
  Hong Kong SAR\\
  \texttt{\{chuanlin.lan, ziyuanyin3-c\}@my.cityu.edu.hk} \\
  \texttt{\{arinbasu, rosachan\}@cityu.edu.hk} \\
}

\begin{document}
\maketitle

\begin{abstract}
3D hand tracking methods based on monocular RGB videos are easily affected by motion blur, while event camera, a sensor with high temporal resolution and dynamic range, is naturally suitable for this task with sparse output and low power consumption. However, obtaining 3D annotations of fast-moving hands is difficult for constructing event-based hand-tracking datasets.
In this paper, we provided an event-based speed adaptive hand tracker (ESAHT) to solve the hand tracking problem based on event camera. We enabled a CNN model trained on a hand tracking dataset with slow motion, which enabled the model to leverage the knowledge of RGB-based hand tracking solutions, to work on fast hand tracking tasks.
To realize our solution, we constructed the first 3D hand tracking dataset captured by an event camera in a real-world environment, figured out two data augment methods to narrow the domain gap between slow and fast motion data, developed a speed adaptive event stream segmentation method to handle hand movements in different moving speeds, and introduced a new event-to-frame representation method adaptive to event streams with different lengths.
Experiments showed that our solution outperformed RGB-based as well as previous event-based solutions in fast hand tracking tasks, and our codes and dataset will be publicly available.

\end{abstract}


\section{Introduction}
\label{sec:intro}

Hand tracking, or hand pose estimation, is a critical topic in the realization of touchless gesture-based human-computer interaction (HCI). With the continuous development of better deep-learning models, most of the current hand tracking algorithms are based on frames recorded by RGB or depth cameras ~\cite{2020Hand}~\cite{2018GANerated}~\cite{2021RGB2Hands}. However, due to the limitations of this hardware with low temporal resolution (usually 30-60 fps), captured images of hands in rapid motion are blurred and limit the performance of hand tracking algorithms.

\begin{figure*}[ht!]
    \centering
    \includegraphics[width=\textwidth]{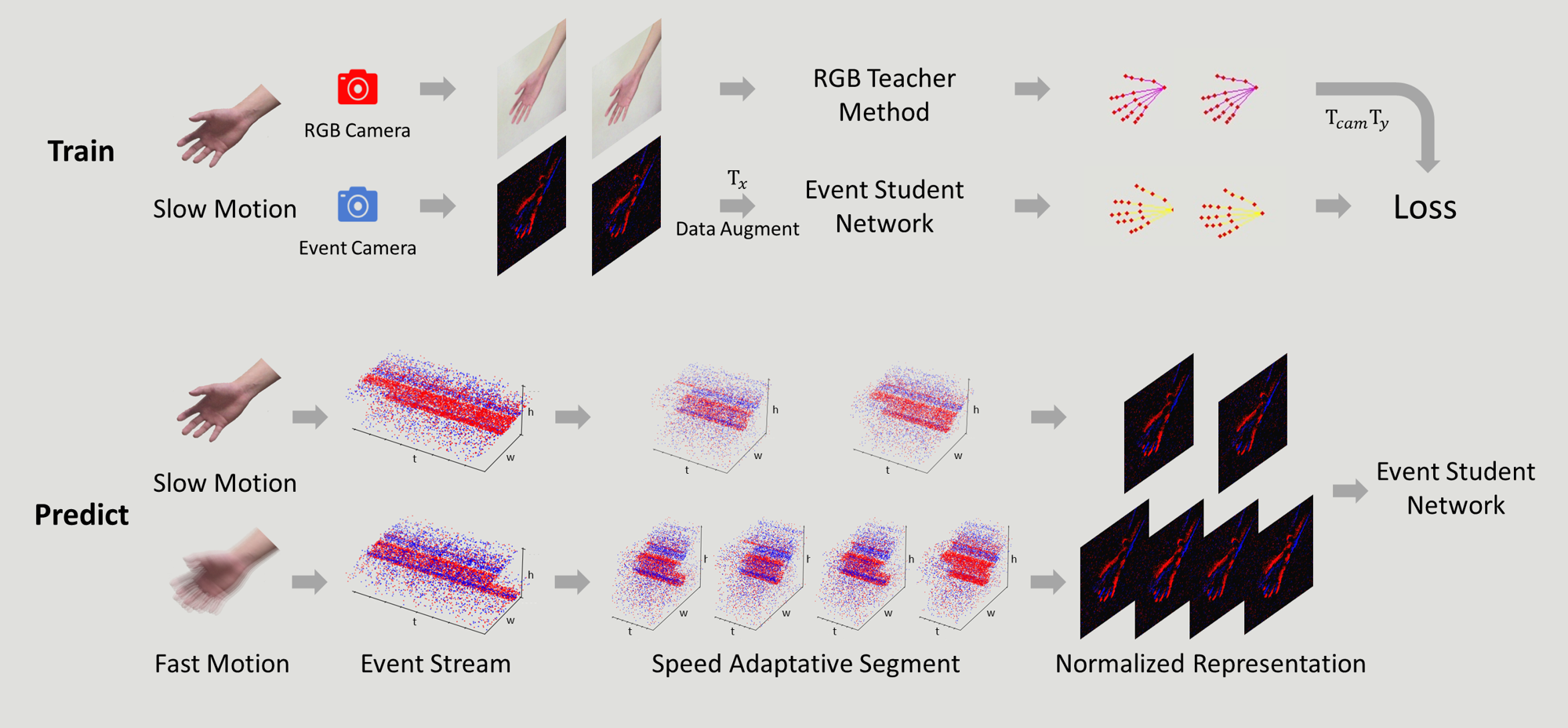}
    \caption{Our fast hand tracking solution is based on the idea \textit{tracking fast by learning from slow} to sidestep the difficulty in obtaining 3D annotations of hands in fast motion. We applied 2 RGB cameras and 1 event camera to record slow hand motion as trainset, trained an event student network with supervising information from RGB-based teacher method, and enabled the event student network to solve speed adaptive hand tracking task in the prediction procedure. To narrow the gap between the trainset of slow motion and testest of fast motion, we figured out two data augment methods, a speed adaptive event stream segmentation method and a event-to-frame representation method.}
    \label{fig:pipeline}
\end{figure*}

One previous possible solution considers RGBD cameras with high frequency. However, RGB cameras with high temporal resolution usually require a high illumination environment, limiting their application scenarios. The large-scale data flow of such RGB cameras also challenges data transmission, storage, and computation system. Depth cameras based on binocular vision have a similar problem, while time-of-flight (ToF) depth sensors with a high frame rate are not available. Another possible solution applies additional data generated by other high-frequency sensors. ~\cite{20203D} proposed a multi-modality method, fusing 100Hz gyroscope data and 30Hz RGBD stream, but the accuracy enhancement brought by gyroscope data was limited. This proposed method is relatively impractical because it also requires extra sensor(s) fixed on hands as well as calibration steps before starting estimating hand pose.

In this work, we utilize an event camera to track hands at different motion speeds. Compared to normal RGB cameras recording frames in a fixed frequency, event cameras could respond to local brightness changes in an asynchronous way, enabling both high temporal resolution and high dynamic range, which makes it optimal hardware to track hands in diverse real-world environments. EventHands ~\cite{2021EventHands} is a pioneer learning-based method to track hand motion in event streams, which constructed a synthetic dataset, developed representation method to transfer event streams to images, and trained a CNN consisting of ResNet18 and MANO model to solve the problem. However, the domain gap between real and synthetic datasets caused a performance drop when validating the model on real data. Consequently, a dataset collected in the real environment is of vital importance for event hand tracking. 

However, obtaining the 3D annotation is a challenging task. Annotation methods like data glove and VICON system would change the appearance of hands, affect the captured event stream and thus are not suitable for this problem. Another opiton is to apply an RGB binocular stereo vision system to get the 3D label of streams recorded by an event camera, which allow us to utilize the knowledge of developed RGB-based method, but give rise to the problem raised at the beginning: how to track hands in fast motion via RGB camera with limited temporal resolution? 

As shown in Fig.~\ref{fig:pipeline}, we sidestepped this problem by applying a deep-learning model trained on a hand tracking dataset with slow motion, which enabled the event hand tracker to leverage the knowledge of RGB-based hand tracking sloution,  to work on fast hand tracking tasks. To narrow the domain gap between slow and fast motion data, 1) in the training procedure we augmented the slow motion data by utilizing event stream of different length to generate event frame as well as randomly suppress the noise in the frame; 2) in the predicting procedure, we figured out a speed adaptive segment method to ensure similar motion range in event frame; 3) we figured out a new representation method, Locally Normalized Event Count and Surface (LNECS), which could preserve the time information and reduce the impact of different noise distribution of event stream of varying length. It is worth noting that the above methodology can be applied not only to fast hand tracking tasks but also to other fast-moving object detection or tracking tasks, especially when the annotations of the fast-moving object are impossible or expensive to obtain. We also construct the first event-based hand tracking dataset collected in real environment to train and test our solution. Experiments showed that our solution outperformed RGB-based solutions in fast hand tracking tasks. In summary, our contributions are:

\begin{itemize}
\item We created an Event-based Speed Adaptive Hand Tracker, a solution that uses a deep learning model trained on a hand tracking dataset with slow motion to perform fast hand tracking tasks. This approach allows us to utilize the knowledge from RGB-based hand tracking solutions for event-based fast hand tracking. The same methodology can also be applied to other fast-moving object detection or tracking tasks.
\item We figured out three methods to reduce the domain gap between the training data for slow motion and the prediction data for fast motion. These methods include data augmentation techniques, the speed adaptive segment method, and the LNECS representation method.
\item We constructed and released the first event-based hand tracking dataset collected in the real world.

\end{itemize}

\begin{figure*}[ht!]
    \centering
    \includegraphics[width=\textwidth]{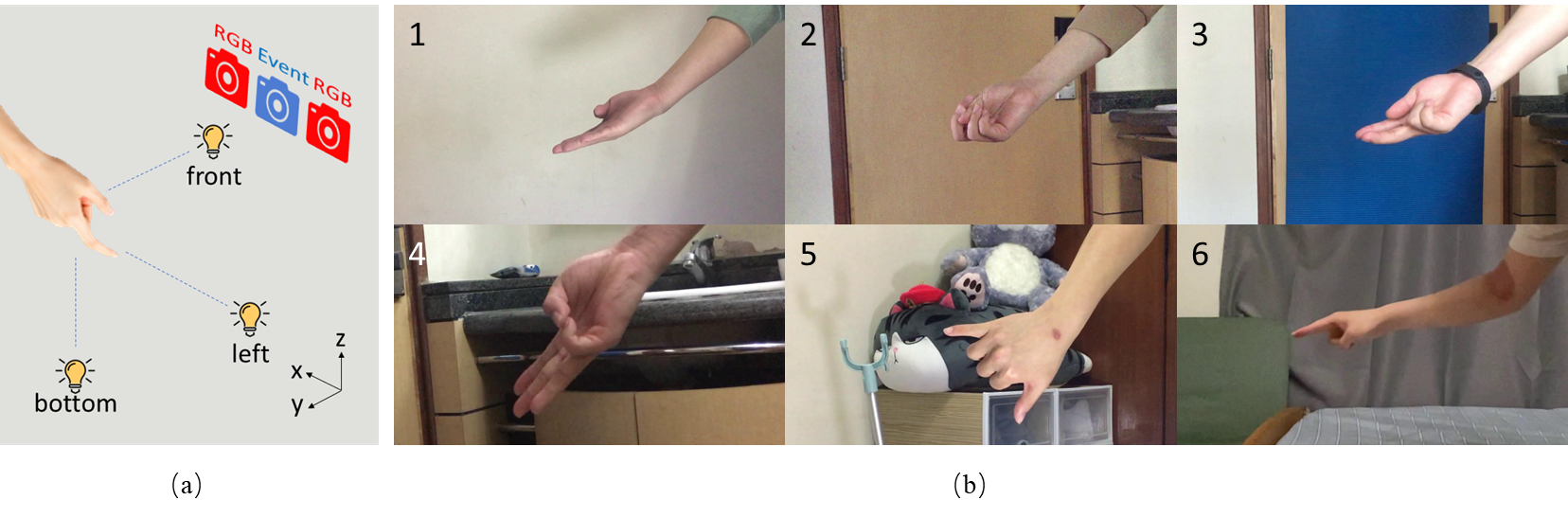}
    \caption{(a): We applied 2 RGB cameras and 1 event camera to collected the dataset with 4 different illumination conditions: without extra light, with extra light from the front/bottom/left side of the hand; (b): The dataset was collected under 6 different indoor backgrounds.We recorded hand with and without an electronic bracelet performing American Sign Language and some random poses to ensure variety of the dataset.}
    \label{fig:dataset}
\end{figure*}

\section{Related Work}
\subsection{3D Hand Tracking Methods}
Most of the existing 3D hand tracking methods were based on RGBD frames, which could be divided into two categories: discriminative model and generative model. Discriminative models estimated hand pose directly from the input frame. To enhance tracking performance, feature fusion operations were widely conducted. Region ensemble network (REN) ~\cite{2017Region} divided the feature map output by a CNN into four parts and fused the features to compute the final result. TriHorn-Net ~\cite{rezaei2022trihorn} first computed the 2D heat map and then fused the 2D feature map, attention map, and depth features to regress the depth value of each joint. Generative models first learned the distributions of the dataset and applied the learned distribution to train a discriminative model. Generalized feedback loop ~\cite{2020Generalized} applied a synthesizer CNN to learn the data distribution and to generate a depth image from the initially predicted hand pose and then utilized an updater CNN to compare the input image and generated image to refine the initial hand pose. All the above-mentioned methods are based on RGBD frames without motion blur, thus lacking the ability to deal with fast-moving hand tracking. 

Due to the differences between the data structures of the event stream and the RGBD frame, the above-mentioned models cannot take the event stream as input directly. Some methods could transfer the event stream to gray frames ~\cite{2021High}~\cite{2021Image}, and SLAM methods could reconstruct depth frames or point clouds from the event stream~\cite{wang2021stereo}. However, the performances of these models could be affected by the domain gap. Extra consumption of bandwidth, computation, and storage without a custom solution for event data would eliminate the advantages of event cameras in real-time processing.

\subsection{Event-based Hand Tracking}
There are several ways to utilize the event stream recorded by the event camera, such as 1) reconstruction of gray images from the event stream, which gives up the temporal information and will cause a huge burden in bandwidth, storage, and computation; 2) conversion of event stream to point cloud, which is not easily understood by the neural network; and 3) event-to-frame representation, which reserves the sparse representation and was adopted in this paper.

Event-to-frame representation includes event occurrence image (EOI), event count image (ECI)~\cite{maqueda2018event}~\cite{zhu2018ev}, surfaces of active events (SAE)~\cite{benosman2013event}, time-surfaces~\cite{lagorce2016hots}, hierarchy of time surfaces ~\cite{lagorce2016hots}, locally-normalized event surfaces (LNES)~\cite{2021EventHands}. The representations adopted in this paper LNECWS was related to ECI and LNES, enabling the model to distinguish hand motion from background noise.

Existing event-based hand tracking methods could be divided into two groups: 1) unsupervised methods like non-rigid 3D tracking, and 2) supervised methods based on synthetic datasets like EventHands~\cite{2021EventHands}. The former method ~\cite{nehvi2021differentiable} deformed a non-rigid model to the desired shape fitting the stuff in the event stream, thus needing accurate initialization and complex computation process, and facing the problem of lacking robustness. The latter reported method employed a synthetic dataset, which was easy to construct compared to the real-world dataset, to train the model, but suffered performance drop in real-world applications due to the distribution gap between real and synthetic datasets.


\section{Our Solution}
To skip annotations for fast hand tracking dataset, the main idea of our solution is to employ a neural networks trained on a slow motion dataset, which allow us to leverage the knowledge of developed RGB-based hand tracking method, to work on fast hand tracking task. The primary challenge of this solution is bridging the domain gap between the training data for slow motion and the prediction data for fast motion. For event segments with the same time length, the motion range in fast and slow hand tracking dataset is different, so a speed adaptive event segment method is needed. In other words, for event segments with similar motion ranges, the time length of event segments in slow and fast hand tracking data will be different, which lead to different noise level as the number of noise events is proportional to the time length. To overcome the domain gap of motion range, we augmented the slow motion data by utilizing event stream of different length to generate event frame and figured out the speed adaptive segment method; to process segments with different time length and noise level, we randomly suppressed the noise in segments of slow motion trainset and created a new representation method, Locally Normalized Event Count and Surface (LNECS).

In this section, we present the details of our real-world dataset (Sec.~\ref{sec:dataset}), as well as the new representation method for handling event segments of varying lengths and noise levels (Sec.~\ref{sec:rep}). We also demonstrate the methods we used to augment the slow motion data and incorporate knowledge from the RGB domain into our training process (Sec.~\ref{sec:tr}) and describe how the trained model track hands at various motion speeds (Sec.~\ref{sec:te}).

\subsection{Dataset}
\label{sec:dataset}
As shown in Fig.~\ref{fig:dataset}, we have considered the following scenarios which are common in real-life hand tracking:
1) Background: We collected the dataset with 6 different indoor backgrounds, which vary from simple to complex.
2) Illumination: Due to the character of the event camera that responds to changes in brightness, different illumination conditions could lead to different outputs of the event camera. We considered 4 illumination conditions while collecting the dataset, which include normal illumination conditions and illumination conditions with extra light from the front/bottom/left side of the hand. Additionally, different illumination conditions would lead to different shadows on the background, which is also challenging for hand tracking algorithms.
3) Self-occlusion and global rotation: Since self-occlusion and global rotation are two major challenges in hand tracking, for all the event streams, we captured poses in American Sign Language (ASL) as well as some random poses to ensure the dataset contains challenging poses with self-occlusion. We also rotated the hand to capture different aspects of each pose.
4) Hand decorations: The decorations on the hand could also affect the performance of hand tracking algorithms. In our dataset, we collected hands with or without an electronic bracelet.

We collected the dataset using an event camera (CeleX5) and two RGB cameras. To ensure the consistency of the data, we calibrated the cameras and matched the timestamp of the event stream and two RGB videos The dataset with slow hand motion contains 40 event streams (5 backgrounds $\times$ 4 illumination conditions $\times$ 2 with or without a bracelet), and dataset with fast hand motion contains 4 event streams (1 background $\times$ 2 illumination conditions $\times$ 2 with or without a bracelet). In total, we recorded about 3.5 hours of event data, and the dataset will be publicly available. 

\subsection{Representation}
\label{sec:rep}

The output of the event camera is a set of events, which could be denoted as $\mathcal{E}=\{e_i\}_{i=0}^N,e_i=(x_i, y_i, p_i, t_i)$, where $e_i$ denotes the event, $x_i$ and $y_i$ denotes the coordinates of the event, and $p_i$ denotes the polarity, and $t_i$ denotes the timestamp. This data format is not suitable for convolutional neural networks, and thus we need to transfer them to frame format. Existing event-to-frame representation methods, such as Event Occur Image (EOI) and Event Count Image (ECI), detected the occurrence or counted the number of the events at each pixel within a time interval, which lost the time information, and some improved methods like, Surfaces of Active Events (SAE) and Locally-Normalized Event Surface (LNES) assigned the value of each pixel as the timestamp of the latest event at the pixel, which, however, are easily affected by noise, especially caused by strobe flash of alternate-current-powered illumination. To solve such problem, we formulated the Locally-Normalized Event Count and Surface (LNECS) based on ECI and LNES. The representation LNECS is denoted with the following formulas:

$${ LNES }(x,y,p) = \frac{max\{t_i|(x_i,y_i,p_i)=(x,y,p)\}-min\{t_i\}} {max\{t_i\}-min\{t_i\}}$$
$${ EC }(x,y,p) = |\{e_i|(x_i,y_i,p_i)=(x,y,p)\}|$$
$${ LNEC }(x,y,p) = \frac{{ EC }(x,y,p)}{max({ EC }(x,y,p))}$$
Then $ { LNECS }$ and $ { LNECWS }$ can be represented as:
$$ { LNECS } = { LNES } \oplus { LNEC }$$
where $\oplus$ denotes concatenate operation in the polarity channel. 

\begin{figure}[t!]
    \centering
    \includegraphics[width=0.4\textwidth]{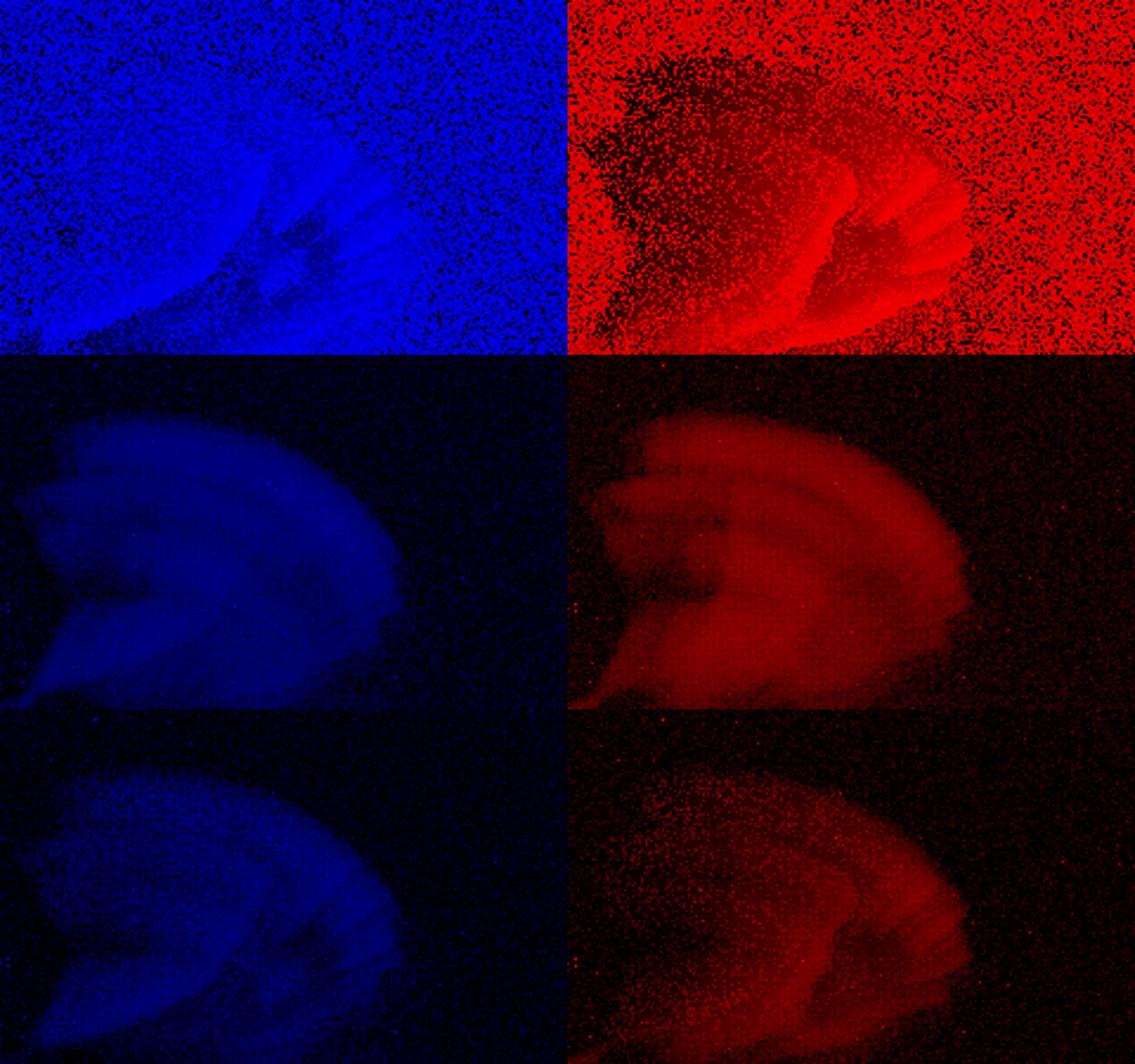}
    \caption{A demo of different representations with two channels. Upper: LNES; middle: LNEC; bottom: LNECWS. }
    \label{fig:repre}
\end{figure}

Figure~\ref{fig:repre} shows a demo of different representations of event streams of a hand moving from left to right. A shown in the upper and middle figure, LNES preserves time information but suffers from noise close to hand and LNEC overcomes the influence of noise but loses time information. Our proposed LNECS combined these representations and integrated both advantages

\begin{table*}[ht!]
\centering
\begin{tabular}{lclcccc}
\toprule
\multirow{2}*{Segment Stardard} & \multirow{2}*{Data Augment} & \multirow{2}*{Representation} & \multicolumn{2}{c}{Fast Motion} & \multicolumn{2}{c}{Slow Motion} \\
& & & AUCp 2D & AUCp 3D & AUCp 2D & AUCp 3D\\
\midrule
10k event number  & No & LNECS   & 0.780 & 0.661 & 0.908 & 0.842\\
\hline
10k event number  & Yes & LNES    & 0.831 & 0.716 & 0.890 & 0.821\\
10k event number  & Yes & LNEC    & 0.837 & 0.717 & 0.897 & 0.828\\
10k event number  & Yes & LNECWS  & 0.836 & 0.722 & 0.891 & 0.822\\
\hline
1k pixel number   & Yes & LNECS   & 0.789 & 0.664 & 0.904 & 0.839\\
2k pixel number   & Yes & LNECS   & 0.801 & 0.674 & 0.912 & 0.843\\
5k pixel number   & Yes & LNECS   & 0.759 & 0.628 & 0.910 & 0.841\\
50ms time length  & Yes & LNECS   & 0.255 & 0.131 & 0.905 & 0.834\\
20ms time length  & Yes & LNECS   & 0.791 & 0.658 & 0.753 & 0.617\\
5k  event number  & Yes & LNECS   & 0.814 & 0.698 & 0.887 & 0.821 \\
20k event number  & Yes & LNECS   & 0.829 & 0.718 & 0.899 & 0.832 \\
50k event number  & Yes & LNECS   & 0.767 & 0.636 & \textbf{0.913} & \textbf{0.845} \\
\hline

event number 10k  & Yes & LNECS   & \textbf{0.840} & \textbf{0.727} & 0.893 & 0.825\\
\bottomrule
\end{tabular}
\caption{Ablation study on slow and fast motion test data. We report the AUCp-2D and the AUCp-3D (higher values are better, bold italic font denotes best numbers).}
\label{tab:abl_seg}
\end{table*}

\subsection{Training Procedure}
\label{sec:tr}
As shown in Fig.~\ref{fig:pipeline}, there are two branches in the training procedure. In the RGB branch, we applied a binocular stereo vision system as the teacher method. An RGB-based hand tracking algorithm initially estimated hand poses in the frames from two RGB cameras, and then, the 3D pose is calculated based on binocular stereo vision algorithm and used as the supervising information of event branch. In the event branch, we augmented the training data to enable the event student network for speed adaptive hand tracking tasks. The parameters of event student network were updated using supervision from the RGB branch after the supervised 3D poses were transformed to event camera space and undergo corresponding data augment transformation. The loss function is calculated by the following function.

$$\mathcal{L} = \Vert f_s ( T_{X} ( x_e))- T_{Y} ( T_{cam} ( f_t(x_r))) \Vert_2$$

where $f_t$ denotes the RGB teacher method, $f_s$ denotes the event student network, $x_e$ denotes the event stream segment, $x_r$ denotes the RGB videos, $T_X$ and $T_Y$ denotes data augment operation and $T_{cam}$ denotes the transform operation from RGB to even camera space.

The data augmentation process includes 3 operations: 1) view augmentation such as rotation and crop, 2) applying event segment of different length, and 3) randomly suppressing noise. The first operation enhances the robustness of the model and the other two operations align the domain of slow and fast hand tracking data. According to the character of event camera, the pixel output an event when detecting brightness changes. A long period of slow motion event segment would be similar to a short period of fast motion event segment regardless the timestamp. Based on this character, we could stimulate fast motion data using longer event segment of slow motion. However, in practice we found that the number of noise events is usually proportional to the time length and thus the noise distribution in augmented data and real fast motion data is different. To align this distribution, we applied random noise suppression augmentation. If the number of events in a pixel and its neighbors is less than a random threshold, the events in this pixel will be regarded as noise and removed. This noise suppression method can be easily implemented by the following functions and required very little computation resource.

$$F'(x,y,p)=1(EC(x,y,p)\ast E_\sigma > \epsilon_r)\times F(x,y,p)$$

where $F(x,y,p)$ and $F'(x,y,p)$ respectively denote the original frame and frame after noise suppression, $E_\sigma$ denotes an average filter kernel with size $\sigma$ and $\epsilon_r$ is a random threshold varying with training sample.

\subsection{Prediction Procedure}
\label{sec:te}
According to the character of event camera, whose pixels respond to local brightness changes independently, fast hand motion would generate more events than slow motion during the same time interval. We utilized this character and figured out our speed adaptive segment method. Instead of dividing event stream into segments with the same time interval, our speed adaptive segment method determine the length of event segments by event number. In this way, we ensure the consistency of motion range in a representation frame despite speed changes.

\begin{table*}[ht!]
  \centering
    \begin{tabular}{cccccccc}
    \toprule
    Solution & Segment Standard &  Representation & \multicolumn{2}{c}{Our Fast Testset}  &  \multicolumn{2}{c}{Our Slow Testset} & EventHands Testset \\
                        &                             &                            & AUCp\_2d       & AUCp\_3d             & AUCp\_2d & AUCp\_3d                   & AUCp\_2d\\
    \midrule
    EventHands  & event number & LNES    & 0.652 & 0.577 & 0.741 & 0.652 & \\
    EventHands  & time length  & LNES    & 0.033 & 0.029 & 0.038 & 0.032 & 0.654\\
    Ours        & event number & LNECS   & \textbf{0.840}  & \textbf{0.727} & 0.893 & 0.825 & 0.665 \\
    Ours        & event number & LNES    & 0.831 & 0.716 & 0.890 & 0.821 & 0.738\\
    Ours        & event number & LNEC    & 0.837 & 0.717 & \textbf{0.897} & \textbf{0.828} & \textbf{0.773}\\
    Ours        & event number & LNECWS  & 0.836 & 0.722 & 0.891 & 0.822 & 0.767\\
    \bottomrule
    \end{tabular}
  \label{tab:eh_result}
  \caption{Comparison between our solution and EventHands.}
\end{table*}

\section{Experiment}
\subsection{Implement Details}

We employed MediaPipe~\cite{zhang2020mediapipe} as the hand tracking model in RGB teacher method and ResNet-18 as the event student network and tested our solution on event streams of slow and fast motion. The resolution of our event camera CeleX5 is $1280 \times 800$ and we generate the frame for the event student network with size $240 \times 150$. We use an Adam optimizer with an initial learning rate of 10-4 and a batch size of 128. We train the model for 50 epochs, reducing the learning rate by a factor of 10 when the validate loss value did not decay in the last 5 epochs. 

We first introduce our evaluation dataset and metrics. Then we present ablation studies to evaluate our design choices and compare our solution against RGB-based solutions. 

\subsection{Test Data and Metrics}
The test data includes a slow motion dataset with 8 slow-motion event streams as well as a fast motion dataset 4 fast-motion event streams. The annotations of slow hand motion were obtained via a 30Hz RGB binocular stereo vision system, and annotations of fast hand motion were obtained via a 240Hz system. Notably, the annotation obtained by 240Hz system aims to validate performance on the fast hand tracking task of models trained on the slow hand tracking dataset, but this does not mean that our solution could only deal with hand motion with speed clear under 240Hz RGB camera. We obtained the 3D ground truth based on initial hand pose estimate from the two RGB videos by MediaPipe~\cite{zhang2020mediapipe}, and manually exclude the inaccurate results.

For metrics, we applied palm-normalized percentage of correct keypoints ($PCK_p$) which defines a candidate keypoint to be correct if it falls within a given distance normalized by palm length around the ground truth. and the area under the $PCK_p$ curve ($AUC_p$) to make the result comparable under different modalities. Specifically, we calculated the distance between the wrist and middle finger MCP annotations as the palm length.

\subsection{Ablation Study}

\subsubsection{Speed Adaptive Segmentation Method}

We compared the results of different segmentation methods using various standards, including the same pixel number standard, the same time length standard, and our proposed same event number standard. Our findings revealed that the performance of our proposed same event number standard was more sensitive to changes in the event number when applied to fast test data. The largest performance difference among slow motion samples with different event numbers was 0.892 and 0.913, while the difference among fast motion samples was much larger, at 0.674 and 0.840. Additionally, the solution achieved its best performance on slow samples with 50,000 events and on fast samples with 10,000 events. Furthermore, the performance difference was minimal when using test data samples with 10,000 events, indicating that the minimal domain gap between slow and fast motion data is reached with fewer events in data samples.

For fixed temporal length segments, we generated a dataset comprising event segments with a length of 20ms and 50ms, on which the model achieved its best performance for fast and slow test data, respectively. However, results showed that the solution using this segmentation standard always suffered a performance drop, indicating that this standard is unable to narrow the domain gap between slow and fast motion data.

The standard of using the same pixel number required generating event frames that contained the same number of pixels where the event occurred. The solution that applied this standard obtained the best performance on slow test data, but was not competitive on fast motion test data. A possible reason for this is the influence of different noise distributions, which leads to a larger motion range in fast motion frames to reach the same pixel number as slow motion frames, which contain more noise pixels. This standard also requires more complex operations during implementation, as it necessitates repeatedly generating event frames and checking pixel numbers, which requires loop and condition operations, while our proposed standard can segment data samples from event stream directly.

\subsubsection{Event-to-Frame Representation}
We compared different event-to-frame representation methods and results showed that our LENCS representation outperformed other representations. Despite the representations introduced in Sec 3.2, we also test the representation Locally-Normalized Event Count Weighted Surface (LNEWCS), which can be denoted as:
$$ { LNEWCS }(x,y,p) = { LNES }(x,y,p) \times { LNEC }(x,y,p) $$
This representation also preserves the time information and avoid the influence of noise as shown in Fig.~\ref{fig:repre}. The reason for the unsatisfying result is it only contains 2 channels, which is the same as LNES and LNEC, while LNECWS contians 4 channels, indicting that the time and event count information cannot be compressed into one channel.

\subsubsection{Data Augmentation}
In Sec.~\ref{sec:tr}, we introduced our data augmentation method. Augmentation does not help on slow motion data since there is no domain gap to be bridged. On fast motion data, using data augmentation significantly improves the quality of the predictions.

\subsection{Comparison with EventHands}
EventHands is the pioneer deep-learning based hand tracking solution applying event camera, which enabled a ResNet-18 trained on synthetic data to deal with fast hand tracking tasks in the real world. 
We utilized the released trained model, test data as well as the evaluation code of EventHands and compared the performance of our solution and EventsHands on both testset. Notably, the backbone in both solutions is ResNet-18 and thus would not affect the comparison. The EventHands divided the event stream into segments of 100ms, but this segmentation methods lead to unsatisfying performance on our test data, so we also tested the performance of the trained model in EventHands with our speed adaptive segmentation method and selected the best result. The EventHands testest was captured with a different event camera, which was less sensitive than our camera and output less events for similar motion range. As a result, our solution obtained the best performance with event segments of 4,000 events on the EventHands testset.

As shown in Table.~\ref{tab:eh_result}, our solution outperformed EventHands in both testset, indicating that the idea of tracking fast by learning from slow is better than the idea of tracking real by learning from synthetic.
Besides, as discussed in Sec~\ref{sec:rep}, LNECS combined the time information and event number information, which enable the deep learning model to learn and distinguish the background noise in the dataset. However, in the testset of EventHands, noise often appeared at the previous position of the hand, which did not appear in our dataset and cannot be stimulated by our data augmentation methods. This gave rise to the most severe performance drop of solutions applying LNECS than other representations. In contrast, solution employing LNEC is less likely to be influenced by noise, and obtained the best performance on EventHands testset.

\begin{table}[t!]
\centering
\begin{tabular}{lcc}
\toprule
Method & AUCp2D & AUCp3D \\
\midrule
Mediapipe & 0.735 & 0.644 \\
Minimal-hand & 0.508 & 0.345 \\
Ours(MNECS) & \textbf{0.840} & \textbf{0.727} \\

\bottomrule
\end{tabular}
\caption{Comparison of performance of our solution and RGB-based solution on fast hand tracking task.}
\label{tab:rgb}
\end{table}

\subsection{Comparison with RGB-based Method}
We also captured 30Hz rgb videos for fast hand tracking test data, whose annotations were obtained in the same way as the way to obtain annotations for event stream. We tested two RGB based State-Of-The-Arts solutions Mediapipe~\cite{zhang2020mediapipe} and Minimal-hand~\cite{zhou2020monocular}.

As shown in Table.~\ref{tab:rgb}, our event based method outperformed RGB based State-Of-The-Arts. Notably, the annotations of fast motion dataset are obtained from 240Hz RGB video processed by Mediapipe, which means that the performance of Mediapipe on slow motion tasks with no blur porblem is 1. The performance drop of RGB-based method is 0.265 on 2D task and 0.356 on 3D task, while the performance drop of our event-based solution is 0.053 on 2D task and 0.098 on 3D task.

\section{Conclusion and Future Work}
We have developed an innovative solution called the Event-based Speed Adaptive Hand Tracker. This system employs a deep learning model that has been trained using a dataset of hand tracking in slow motion, enabling it to accurately track fast-moving hands. This approach allows us to leverage the knowledge gained from RGB-based hand tracking technique and the same methodology can also be adapted for other types of fast-moving object detection or tracking.

To achieve this, we employed three methods to bridge the gap between the training data for slow motion and the prediction data for fast motion. These methods include data augmentation, speed adaptive segmentation and NECS representation. Additionally, we have created and released the first-ever event-based hand tracking dataset that was collected in real-world environments. Results showed that our proposed solution enhanced the performance in fast hand tracking tasks, and outperformed other RGB-based methods as well as previous event-based method.

For future work, our current method only applied slow motion data in the training procedure, and utilizing unlabeled fast motion data and domain adaptive method would be a possible way to further improve the method. Besides, the performance of our method is limited by the performance of applied RGB model and adding synthetic data in the training process may also help to enhance the prediction accuracy.

\bibliographystyle{unsrt}  
\bibliography{references}

\end{document}